\documentclass[10pt,twocolumn,letterpaper]{article}

\usepackage{cvpr}              
\definecolor{cvprblue}{rgb}{0.21,0.49,0.74}
\usepackage[pagebackref,breaklinks,colorlinks,allcolors=cvprblue]{hyperref}
\usepackage{tabularray}
\usepackage{graphicx}
\usepackage{float}
\usepackage{algorithm}
\usepackage{algorithmic}
\usepackage{fontawesome5}
\def\paperID{****} 
\def\confName{CVPR}
\def\confYear{2026}

\title{Accelerating Diffusion-based Video Editing via Heterogeneous Caching: \\ Beyond Full Computing at Sampled Denoising Timestep}

\author{
  Tianyi Liu\textsuperscript{1}\quad\quad\quad
  Ye Lu\textsuperscript{1}\quad\quad\quad
  Linfeng Zhang\textsuperscript{2}\quad\quad\quad
  Chen Cai\textsuperscript{1}\quad\quad\quad
  Jianjun Gao\textsuperscript{1}\\
  Yi Wang\textsuperscript{3}\quad\quad\quad
  Kim-Hui Yap\textsuperscript{1~\faEnvelope[regular]}\quad\quad\quad
  Lap-Pui Chau\textsuperscript{3}\\[4pt]
  \normalsize{\textsuperscript{1}Nanyang Technological University\quad
  \textsuperscript{2}Shanghai Jiao Tong University\quad
  \textsuperscript{3}The Hong Kong Polytechnic University}\\
  {\tt\footnotesize \{liut0038, lu0001ye, e190210, gaoj0018\}@e.ntu.edu.sg}\quad
  {\tt\footnotesize zhanglinfeng@sjtu.edu.cn}\\
  {\tt\footnotesize ekhyap@ntu.edu.sg}\quad
  {\tt\footnotesize \{yi-eie.wang, lap-pui.chau\}@polyu.edu.hk}
}

\begin{document}
\maketitle


\begin{abstract}
Diffusion-based video editing has emerged as an important paradigm for high-quality and flexible content generation. 
However, despite their generality and strong modeling capacity, Diffusion Transformers (DiT) remain computationally expensive due to the iterative denoising process, posing challenges for practical deployment. 
Existing video diffusion acceleration methods primarily exploit denoising timestep-level feature reuse, which mitigates the redundancy in denoising process, but overlooks the architectural redundancy within the DiT that many attention operations over spatio-temporal tokens are redundantly executed, offering little to no incremental contribution to the model’s output.
This work introduces HetCache, a training-free diffusion acceleration framework designed to exploit the inherent heterogeneity in diffusion-based masked video-to-video (MV2V) generation and editing. 
Instead of uniformly reuse or randomly sampling tokens, HetCache assesses the contextual relevance and interaction strength among various types of tokens in designated computing steps. 
Guided by spatial priors, it divides the spatial-temporal tokens in DiT model into context and generative tokens, and selectively caches the context tokens that exhibit the strongest correlation and most representative semantics with generative ones. 
This strategy reduces redundant attention operations while maintaining editing consistency and fidelity. 
Experiments show that HetCache achieves a noticeable acceleration, including a 2.67$\times$ latency speedup and FLOPs reduction over commonly used foundation models, with negligible degradation in editing quality.
\end{abstract}    
\vspace{-1em}
\section{Introduction}
\label{sec:intro}

\begin{figure}[t]
    \centering\includegraphics[width=\linewidth]{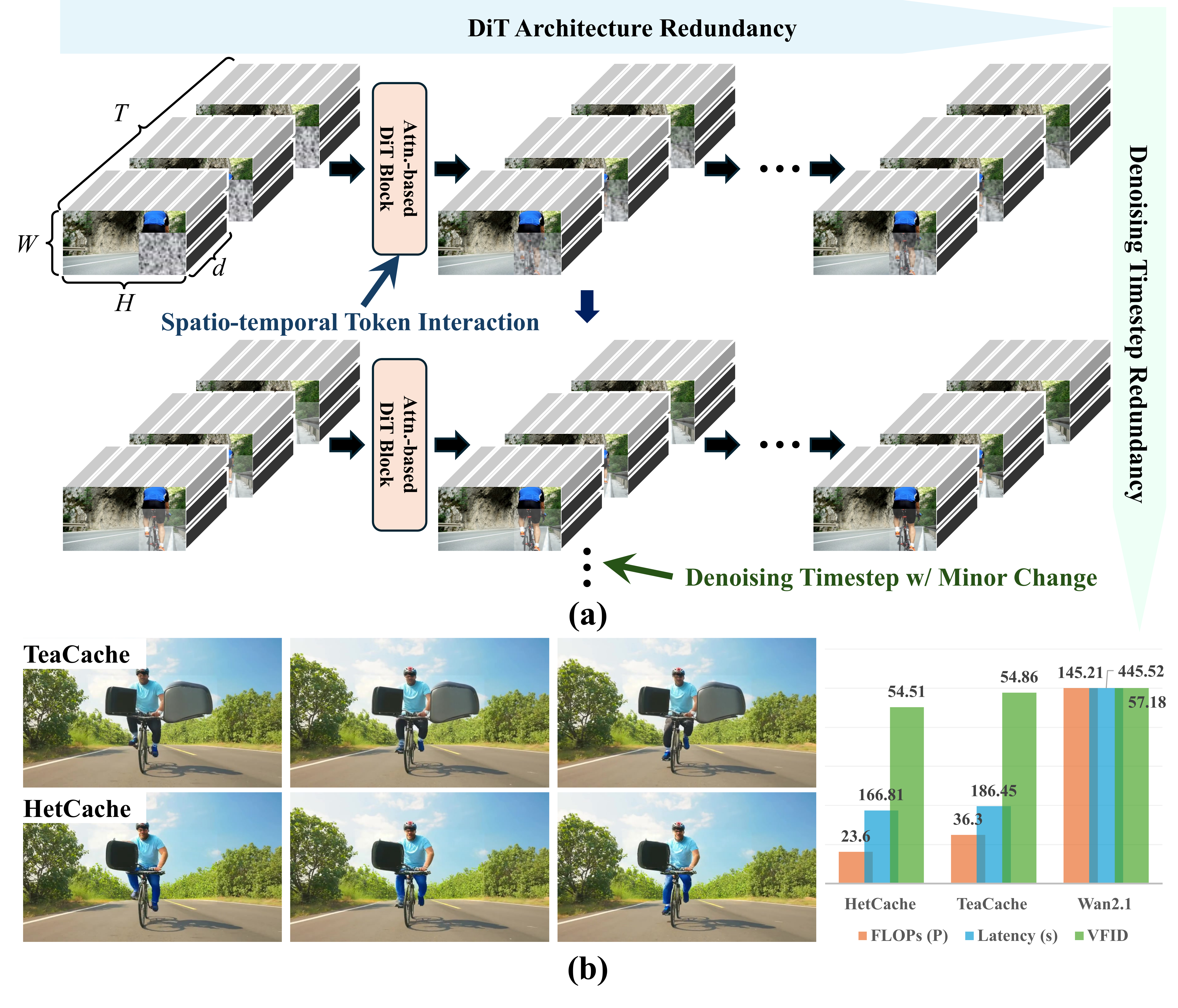}
    \vspace{-2.5em}
    \centering\caption{\textbf{(a).} Illustration of the acceleration dimensions in Diffusion Transformers (DiTs). Unlike existing methods, the proposed Heterogeneous Caching (HetCache) jointly models denoising-step redundancy in the diffusion process and token redundancy within the Transformer backbone. \textbf{(b).} As a tailored heterogeneous strategy, HetCache accelerates diffusion-based masked video-to-video (MV2V) editing while maintaining generation quality.}
    \vspace{-2em}
    \label{fig:teaser}
\end{figure}

Diffusion-based generative methods have recently gained attention in various video editing tasks~\cite{wan2025wan}. 
With Diffusion Transformers (DiTs), which adopt the Transformer as the denoising backbone, both visual quality and generalization ability have been significantly improved in video synthesis and editing~\cite{ho2022video, ma2024latte}. 
Through scalable parameterization, DiTs provide larger modeling capacity and finer spatio-temporal representations, enabling more flexible generation across complex scenes~\cite{peebles2023scalable}. 
However, these advantages come with substantial computational cost. 
The dense interactions among spatio-temporal tokens within the Transformer layers lead to high computational complexity, while the iterative nature of the denoising process in diffusion models requires repeated network forward evaluation over multiple timesteps, resulting in significant inference latency. 
These factors collectively constrain the real-time and interactive applications for diffusion-based video editing.

As the demand for efficient and lightweight video generation continues to grow, recent studies on accelerating video diffusion models have explored knowledge distillation~\cite{meng2023distillation} and post-training optimization methods~\cite{nitzan2024lazy}. 
These approaches typically rely on knowledge transfer between teacher and student models or weight quantization to reduce inference cost, but they inevitably require additional training and data resources, resulting in increased computational overhead. 
To eliminate retraining costs, subsequent work has investigated training-free acceleration strategies, among which feature caching has gained particular attention. 
By caching and reusing intermediate features in denoising timesteps, such methods achieve acceleration within the diffusion framework without modifying model parameters~\cite{zou2025accelerating, liu2025timestep, zhao2025realtime}. 
However, existing approaches primarily focus on temporal redundancy across timesteps, while neglecting the token redundancy and heterogeneity introduced by the viideo editing task and additional temporal dimension in Transformer-based video models. 
This omission limits both the flexibility and the upper bound of current caching-based acceleration schemes.

In this work, we aim to develop a more efficient feature caching mechanism for video diffusion models tailored to masked video-to-video (MV2V) generation and editing tasks. 
Our design considers two complementary sources of redundancy including 1) timestep-level redundancy in the denoising process, and 2) spatio-temporal token redundancy within the attention layers of Diffusion Transformers (DiTs). 
The main challenge lies in identifying redundant tokens in a video editing context where spatial and temporal dependencies are highly uneven.
Unlike general video generation, the MV2V task setting features explicit regions of interest (ROI)~\cite{jiang2025vace}. 
Therefore, applying uniform caching to all tokens within a denoising step for computation can degrade the reconstruction quality inside the masked area. 
Intuitively, the attention mechanism should allow a minimal number of context(unmasked) tokens to provide strong semantic guidance for the generative(masked) tokens, ensuring sufficient representation quality while reducing computational cost. 
However, the representational importance and interaction strength of context tokens are only observable after attention computation. 
Without an effective mechanism to estimate these properties beforehand, token sampling may negatively affect the quality of generated content.

To address this problem, we propose Heterogeneous Caching (HetCache), a training-free caching strategy designed for efficient inference of video editing. 
The key idea of HetCache is that both denoising timesteps and context tokens in Diffusion Transformers (DiTs) contribute unequally to the final generation quality. 
By modeling this heterogeneity across temporal and token dimensions, HetCache performs selective caching that adapts to each dimension independently.
During inference, HetCache first identifies anchor timesteps where model output is expected to change significantly and performs full computation at these steps.
Within each anchor timestep, unmasked tokens are divided into two groups based on spatial priors: context tokens, which are subject to selection, and margin tokens, which are fully preserved around the masked boundary. 
These tokens are further clustered in the semantic space, and the attention interactions between context tokens and masked generative tokens are then analyzed to estimate their semantic relevance, allowing the model to identify informative context tokens. 
In subsequent timesteps, the cached representative tokens replace the full set of context tokens during the attention computation, forming partial computing steps. 
This design effectively reduces the number of active tokens without compromising generation fidelity, thus achieving acceleration for diffusion-based video editing.

The contributions of this work are summarized below.

\begin{itemize}
\item \textbf{Token analysis for diffusion-based video editing.}
We analyze the token-wise redundancy in DiT-based MV2V generation and editing, revealing the inherent token heterogeneity caused by the region-of-interest (ROI) nature.
\item \textbf{A token-level caching mechanism for efficient diffusion-based video editing.}
We propose HetCache, a training-free caching framework that performs heterogeneous caching across both denoising timesteps and spatio-temporal tokens. 
By adapting caching and reuse strategies to the characteristics of each dimension, HetCache introduces partial denoising steps guided by expected output variation and reduces the attention computation through semantic representativeness and interaction-based selection.
\item \textbf{Comprehensive evaluation and state-of-the-art efficiency.}
Extensive experiments and evaluation using common DiT backbones for video completion and text-guided MV2V editing on VACE-Benchmark and VPBench demonstrate that HetCache achieves an improved balance between generation quality and computational efficiency, providing a practical solution toward real-time and interactive diffusion-based video editing.
\end{itemize}
\section{Related Works}\label{sec:related_works}

\subsection{Diffusion-based Video Editing}



Diffusion models have evolved from U-Net backbones~\cite{ho2020denoising,nichol2021improved} to Diffusion Transformers (DiTs)~\cite{peebles2023scalable,chen2024gentron,li2024efficient}, improving scalability and generation quality, but also increasing inference cost. 
In practice, representative DiT systems report consistent quality gains at the cost of higher per-step compute, which amplifies the latency bottleneck under many sampling steps.
In recent years, diffusion-based video editing can be viewed as a conditional video-to-video (V2V) generation problem~\cite{liang2023flowvid} (often “MV2V”) with explicit guidance such as text prompts, spatial masks (ROI), or structural hints (e.g., depth/optical flow). 
Canonical applications include inpainting~\cite{xie2023smartbrush}, object removal/replacement~\cite{wang2023instructedit}, and stylization~\cite{huang2024diffstyler}; recent unified pipelines (e.g., “all-in-one” creation/editing) integrate multiple controls in the diffusion loop~\cite{hu2022unified}. 
Compared with unconditional generation, editing stresses accurate propagation of edits within the ROI while preserving consistency elsewhere, which makes token-level interactions around masks especially critical.

\subsection{Diffusion Model Acceleration}

\noindent \textbf{Architectural Optimization. }Two common directions reduce the denoiser’s cost: (i) parameter-centric compression—structured/unstructured pruning~\cite{fang2023structural,fang2023depgraph} and post-training quantization—to shrink compute/memory~\cite{wang2024towards}, and (ii) token/path-centric efficiency—module or token-sequence simplification~\cite{kim2024token} (e.g., token merging/pruning) to lower attention/MLP load. Although effective, these methods typically require fine-tuning or calibration and introduce non-trivial engineering overhead.

\noindent \textbf{Training-free Acceleration. }Training-free methods avoid re-training and fall into two families. (a) Sampler acceleration lowers the number of denoising steps via deterministic samplers or high-order ODE solvers~\cite{salimans2022progressive}; step distillation/consistency further compresses steps but may trade off fidelity at low step counts~\cite{li2023autodiffusion}. (b) Feature caching reduces redundant compute by reusing intermediate features across timesteps~\cite{zou2024accelerating}. For U-Net denoisers, cache-and-reuse along skip/encoder paths achieves notable speedups~\cite{wimbauer2024cache}. For DiTs, recent works extend caching to Transformer blocks~\cite{qiu2025gradientoptimizedcache,liu2025fastcache} (e.g., caching features or residuals, pyramid broadcast for video). However, most DiT accelerators apply homogeneous cache decisions to all tokens inside a timestep. More recent analyses~\cite{liu2025timestep} highlight that tokens differ in temporal redundancy and error propagation sensitivity; token-wise caching in DiTs~\cite{zou2025accelerating} therefore selects which tokens to cache and where to reduce attention and MLP workload with smaller quality loss.

For video diffusion transformers under MV2V tasks, ROI-induced spatio-temporal heterogeneity makes uniform per-timestep cache/prune choices sub-optimal: context tokens outside the mask should provide strong but sparse guidance~\cite{nitzan2024lazy}, while masked tokens require full updates to maintain edit fidelity. 
This motivates heterogeneous, editing-aware caching that couples 1) timestep selection and 2) token-level selection tailored to editing task.
\section{Method}

\subsection{Preliminaries}
\noindent \textbf{Diffusion Models.}
Diffusion models~\cite{ho2020denoising} are generative models that synthesize data by learning to reverse a gradual noising process. 
Given a clean image $x_0$ sampled from a real data distribution, the forward process progressively adds Gaussian noise over $T$ timesteps with a noise schedule ${\alpha_t}*{t=1}^T$, which monotonically decreases with $t$, ensuring a smooth transition from data to noise.
After $T$ steps, $x_T$ approximates pure Gaussian noise.
The reverse process learns to denoise $x_t$ step by step via a neural network $\epsilon*\theta(x_t, t)$ that predicts the added noise.

Traditionally, U-Net architectures have been widely adopted to model $\epsilon_\theta$ and have achieved strong generation quality. However, recent research demonstrates that transformer-based backbones exhibit superior scalability and global reasoning ability, giving rise to the Diffusion Transformer (DiT) family.
DiT~\cite{peebles2023scalable} replace the convolutional U-Net backbone with a fully transformer-based architecture, achieving state-of-the-art performance across image and video generation tasks. Given an input feature map $x_t$, it is reshaped into a sequence of tokens ${x_i}*{i=1}^{H \times W}$, each representing a spatial patch of the image. The denoising network can be formulated as a stack of transformer blocks $\mathcal{G} = g_1 \circ g_2 \circ \dots \circ g_L$, where each block $g_l$ consists of self-attention ($f_\mathrm{SA}^l$), optional cross-attention ($f_\mathrm{CA}^l$) for conditional generation, and a feed-forward network ($f_\mathrm{MLP}^l$).
Timestep embeddings and, when applicable, text embeddings are injected into each block via adaptive normalization or cross-attention, guiding the denoising trajectory. 
The transformer-based formulation enables large-scale modeling, long-range dependency learning, and unified applicability to diverse generative tasks such as text-to-image, image-to-video, and text-to-video synthesis.

\subsection{Heterogeneity Investigation}
\begin{figure*}[t]
    \centering\includegraphics[width=1\linewidth]{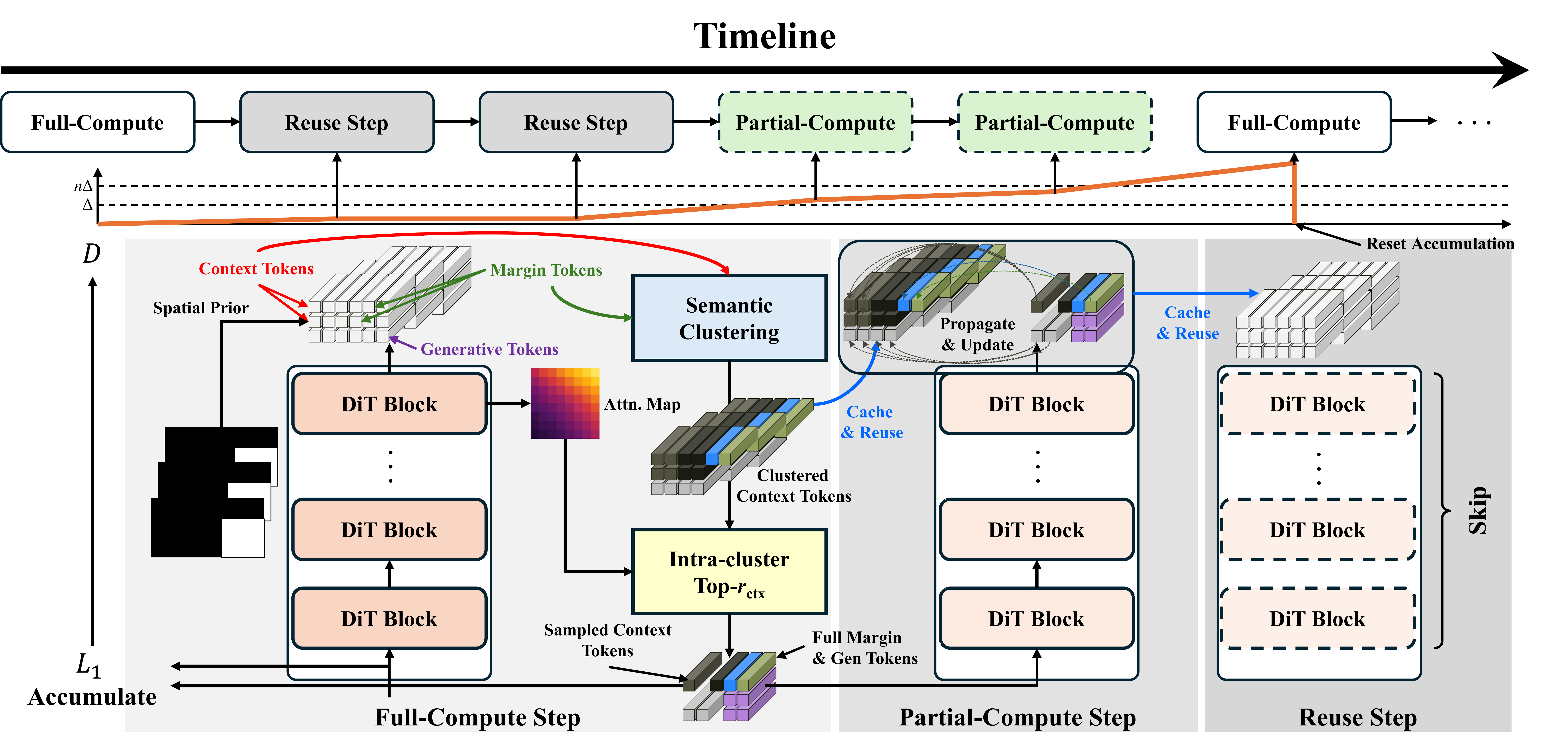}
    \centering\caption{The overview of our proposed HetCache scheme. In denoising process, we use the timestep-embeddings-modulated-input~\cite{liu2025timestep} to estimate the computing demand. According to the accumulated distance, \textbf{Full-Compute} anchor step, \textbf{Reuse} step and \textbf{Partial-Compute} step will be executed. In full-computing, HetCache will use spatial prior extracted from editing mask to categorize the DiT tokens into \textbf{\textcolor{red}{Context}}, \textbf{\textcolor{ForestGreen}{Margin}}, and \textbf{\textcolor{violet}{Generative}} Tokens. The Context Tokens which takes high portion and cause redundant computation cost will be cached for partial-compute steps according to its semantic representativeness and interaction strength with the generative tokens.}
    \label{fig:overview}
    \vspace{-1em}
\end{figure*}

\noindent \textbf{Spatiotemporal Heterogeneity.}
Diffusion-based MV2V generation and editing inherently exhibits spatio-temporal heterogeneity during the denoising process.
Instead of performing a uniform global refinement across timesteps, it has been discussed for video generation that the denoising dynamics vary significantly over time and across regions.
Early timesteps tend to reconstruct coarse structural layouts, whereas later ones refine high-frequency details. 
Even within a single timestep, spatial regions evolve asynchronously—motion-dominant or masked areas often change faster or slower than static backgrounds~\cite{ho2022video}.
This indicates that the diffusion process is not temporally uniform or spatially synchronized; rather, it progresses in a level-adaptive manner modulated by both timestep embeddings and content dynamics. 

\noindent \textbf{ROI-driven Token Interaction.}
In addition to the heterogeneity in the timestep dimension, for MV2V editing, the ROI nature determines that the interaction between context tokens (unmasked) and generative tokens (masked) is the core of Transformer inference, which is also emphasized in traditional video editing tasks~\cite{li2022towards, liu2023bitstream, liu2025towards}.
MV2V editing usually focuses on localized modifications, the essential generative behavior arises from the flow of visual and motion information from unmasked regions toward requiring synthesis.
Within DiTs, this interaction is realized through attention layers, where context tokens provide structural guidance while generative tokens reconstruct missing content.
Different tokens exhibit highly unequal sensitivities to attention propagation—errors or updates in certain tokens may spread, while others remain localized~\cite{jiang2025vace}.
Therefore, modeling and selectively enhancing interaction between context and generative tokens is critical to maintaining spatio-temporal coherence in MV2V editing. 

Such multi-dimensional heterogeneity raises the natural argument that existing video diffusion caching strategies overlook the unequal importance of refining timesteps and varied token properties, suggesting that only a subset of denoising steps may require full updates, while others can be partially executed with limited quality loss~\cite{liu2025timestep, zhao2025realtime}. 
This motivates us to analyze DiT in video editing tasks and uncover token-level heterogeneity—differences in temporal redundancy, error propagation, and layer sensitivity to better leverage the heterogeneity for enhanced feature caching. 

\subsection{Caching by Context and Correlation}
The token-level redundancy within a single timestep of DiT enables potential computation reduction. 
However, traditional methods do not exploit it, which motivates our ROI-aware selective caching.
Based on our observation of timestep-level heterogeneity, we categorize denoising timesteps into full-compute steps, partial-compute steps, and reuse steps, allowing us to exploit non-uniform temporal redundancy for efficient caching and more lightweight refinement.
Specifically, following the idea that timestep embedding-modulated noisy inputs correlate strongly with model output variation~\cite{liu2025timestep}, we first compute a per-step difference using the modulated input $F_t = T_t \odot x_t$ as
\begin{equation}
    L_{1}^{\text{rel}}(F,t)=\frac{|F_t - F_{t+1}|_1}{|F_{t+1}|_1},
\end{equation}
where $x_t$ is the latent noise in timestep $t$, $T_t$ is the pretrained timestep embedding, and $\odot$ denotes the modulation.
The relative input change between two adjacent timesteps can be used as a lightweight proxy to estimate output variation.
We then accumulate this difference over consecutive timesteps:
\begin{equation}
D_{a\rightarrow b}=\sum_{t=a}^{b-1} L_{1}^{\text{rel}}(F,t),
\end{equation}
and use the accumulated value $D_{a\rightarrow b}$ to determine the mode of computation of the timestep $b$.

Intuitively, a small accumulated difference indicates that the denoising trajectory is locally stable and can safely reuse cached outputs; a moderate accumulated difference indicates partial drift that benefits from a lightweight refresh; and a large accumulated difference signals significant changes that require full recomputation.
Accordingly, given a cache threshold $\Delta$, we assign each timestep to one of the following regimes: 1) \textbf{Full-compute} step with cache update when $D_{a\rightarrow b} > 1.5\Delta$ and it will perform a full forward pass and full cache refresh.
2) \textbf{Partial-compute} step with EMA-style cache update when $1\Delta < D_{a\rightarrow b} \le 1.5\Delta$ in which only a subset of operations or tokens is recomputed, while cached representations are softly updated.
3) \textbf{Reuse} step when $D_{a\rightarrow b} \le 1\Delta$, in which the cached outputs are reused without recomputation.
This multi-regime scheduling enables fine-grained timestep-level acceleration, where expensive full computations are reserved for moments of high variation, while stable regions of the denoising trajectory benefit from aggressive reuse.

Additionally, guided by the ROI characteristics of video editing, we reorganize the spatio-temporal tokens of DiT during each full-compute step based on their spatial relationship to the editing mask. 
The tokens are partitioned into 
1) \textbf{Context tokens} of unmasked regions far from the edited area, providing global semantic coherence and long-range structural consistency for the generative process.
2) \textbf{Margin tokens} for unmasked tokens adjacent to the mask boundary, directly governing boundary smoothness, geometric continuity, and local blending.
3) \textbf{Generative tokens} representing masked regions that must be synthesized and form the core of the editing operation.
In MV2V generation and editing, these token groups contribute differently: generative tokens define the new content, margin tokens ensure smooth transitions around boundaries, while context tokens are essential for maintaining semantic alignment between the generated region and the rest of the scene.

From a computational standpoint, however, self-attention in DiTs scales quadratically with the number of tokens.
Given $X = h \times w \times t$ total tokens, with $X_c, X_m, X_g$ denoting the counts of context, margin, and generative tokens, the attention cost can be expressed as:
\begin{equation}
    \mathcal{O}(X^2)=\mathcal{O}\big((X_c + X_m + X_g)^2\big).
\end{equation}
While context tokens are semantically crucial, the majority of context-context attention contributes little to the final editing outcome. 
The most critical interactions are 1) the generative-margin interaction, which determines reconstruction fidelity and boundary smoothness, and 2) the generative-context interaction, which enforces semantic consistency, but not the dense context-context interactions that dominate the quadratic cost.

\begin{algorithm}[t]
\small
\caption{HetCache: Caching by Context and Correlation for MV2V Generation and Editing}
\label{alg:hetcache}
\begin{algorithmic}[1]
\STATE \textbf{Input:} model $f_\theta$, timesteps $\{t_T\!\dots t_1\}$, latent $x_T$, mask $M$, thresholds $\tau_{\text{reuse}}=1.5\Delta, \tau_{\text{partial}}=\Delta$, cluster number $K$, selection ratio $r_{\text{ctx}} \in (0,1]$, EMA factor $\gamma$.
\STATE \textbf{Output:} $x_0$.
\STATE Initialize cache $O_{\text{cache}} \leftarrow \varnothing$, cumulative distance $D \leftarrow 0$.
\FOR{$t=T,\dots,1$}
    \STATE Compute modulated input $F_t=T_t\odot x_t$; update $D$ using $d_t=\|F_t-F_{t+1}\|_1/\|F_{t+1}\|_1$ if $t<T$.
    
    \IF{$O_{\text{cache}}\neq\varnothing$ \AND $D\le\tau_{\text{reuse}}$}
        \STATE $O_t\leftarrow O_{\text{cache}}$.
        
    \ELSIF{$D\le\tau_{\text{partial}}$}
        \STATE Split tokens into $\mathcal{X}_{ctx}$, $\mathcal{X}_{mar}$, $\mathcal{X}_{gen}$ via mask $M$.
        \STATE K-Means cluster $\mathcal{X}_{ctx}$ into $\{S_k\}_{k=1}^{K}$ and compute importance $\alpha_i$ from cached $A_{ctx\rightarrow gen}$.
        \STATE Select $\mathcal{X}^{\star}_{ctx}$ by taking top-$r_{\text{ctx}}$ tokens per cluster.
        \STATE Run $f_\theta$ on $\mathcal{X}_{gen}\cup\mathcal{X}_{mar}\cup\mathcal{X}^{\star}_{ctx}$ to obtain $O_t$.
        \STATE $O_{\text{cache}}\leftarrow(1-\gamma)\,O_{\text{cache}}+\gamma\,O_t$; $D\leftarrow0$.
        
    \ELSE
        \STATE Run $f_\theta$ on all tokens to obtain $O_t$.
        \STATE $O_{\text{cache}}\leftarrow O_t$; $D\leftarrow0$.
    \ENDIF
    
    \STATE Update $x_{t-1}$ using $O_t$.
\ENDFOR
\STATE \textbf{return} $x_0$.
\end{algorithmic}
\end{algorithm}

Therefore, our goal is not to weaken the role of context, but to compute it more selectively by preserving only semantically representative and generation-relevant context tokens and ensure full attention fidelity for generative and margin tokens so that we can reduce redundant computations for context-context interaction while retaining necessary semantic guidance.
This design preserves the semantic value of context tokens while effectively reducing computational overhead.
During each \emph{partial-compute step}, we reduce the computational cost of DiT by selecting only semantically representative context tokens for attention computation.
Given the context token set $\mathcal{X}_{ctx}=\{x_i\}_{i=1}^{X_l}$, we perform lightweight K-Means clustering to obtain $\mathcal{S}=\{S_1,S_2,\ldots,S_K\}$ as a semantic partition where the centroid of each cluster is
\begin{equation}
    \mu_k=\frac{1}{|S_k|}\sum_{x_i\in S_k}x_i.
\end{equation}

For each cluster, we estimate the token importance using the cached sparse context-to-generative attention score as 
\begin{equation}
    \alpha_i=\frac{1}{|\mathcal{X}_{gen}|}\sum_{j\in\mathcal{X}_{gen}}\bar{A}_{i,j},
\end{equation}
where $\bar{A}_{i,j}$ aggregates (normalized) attention from context token $i$ to generative token $j$, larger $\alpha_i$ indicates stronger context $\!\rightarrow\!$ ROI contribution.
Then we select the top-$r_{\text{ctx}}$ proportion within each cluster to form the representative set $\mathcal{X}^{\star}_{ctx}$.
This reduces the number of context tokens participating in attention from $X_l$ to $r_{\text{ctx}}X_l$, effectively lowering the attention complexity from
$\mathcal{O}((X_l+X_m+X_n)^2)$
to
$\mathcal{O}((r_{\text{ctx}}X_l + X_m + X_n)^2)$
with minimal overhead, as clustering is performed once per partial-compute step. The overall algorithm is summarized in ~\ref{alg:hetcache}

\begin{table*}[t]
\centering
\scriptsize
\caption{Quantitative evaluation of inference efficiency and visual quality in video generation models. HetCahce achieves superior efficiency and better visual quality across different base models, sampling schedulers, video resolutions, and lengths.}
\label{tab:efficiency}
\resizebox{1\linewidth}{!}{
\begin{tblr}{
  cell{1}{1} = {c=9}{c},
  cell{2}{1} = {r=2}{c},
  cell{2}{2} = {r=2}{c},
  cell{2}{3} = {c=3}{c},
  cell{2}{6} = {c=4}{c},
  cell{2-12}{2-9} = {c},
  cell{1}{1} = {c=9}{c},
  cell{13}{1} = {c=9}{c},
  cell{14}{1} = {r=2}{c},
  cell{14}{2} = {r=2}{c},
  cell{14}{3} = {c=3}{c},
  cell{14}{6} = {c=4}{c},
  cell{13-21}{2-9} = {c},
  vline{2-3,6} = {-}{},
  hline{1,13,22} = {-}{0.2em},
  hline{1,13,22} = {3}{-}{},
  hline{2,14} = {-}{0.1em},
  hline{3,4,5,15,16,17} = {-}{},
  row{1,4,13,16} = {bg=gray!10},
}
{\textbf{Video Inpainting on VACE-Benchmark}}                 &                 &            &                  &           &                           &      &       &                      \\
{\textbf{Method}}                           & {\textbf{Step}}   & {\textbf{Efficiency}}  &       &           & {\textbf{Visual Quality}}   &      &       &                 \\
                                            &                   & {\textbf{FLOPs (P) $\downarrow$}} & {\textbf{Latency (s) $\downarrow$}} & {\textbf{Speed $\uparrow$}} & {\textbf{PSNR $\uparrow$}} & {\textbf{SSIM $\uparrow$}} & {\textbf{VFID $\downarrow$}} & {\textbf{VBench (\%) $\uparrow$}} \\
Wan2.1-VACE                                 & 100               & 145.21          & 445.52          &  $1.00\times$         & 16.06                   & 0.56                & 57.18                & 76.54              \\
Timestep Reduction                          & 50                & 72.60           & 238.84          &  $1.86\times$         & 16.46                   & 0.56                & 54.96                & 76.78              \\
PAB                                         & 50                & 43.56           & 223.20          &  $1.99\times$         & 16.46                   & 0.56                & 56.04                & 76.73              \\
AdaCache                                    & 50                & 39.93           & 242.22          &  $1.83\times$         & 16.46                   & 0.56                & 54.96                & 76.78              \\
FastCache                                   & 50                & 43.56           & 239.10          &  $1.86\times$         & 15.95                   & 0.54                & 68.55                & 71.30              \\
TeaCache - slow                             & 50                & 47.19           & 224.53          &  $2.38\times$         & 16.48                   & 0.56                & 55.49                & 76.43              \\
TeaCache - fast                             & 50                & 36.30           & 186.45          &  $2.53\times$         & 16.51                   & 0.56                & 54.86                & \textbf{76.80}     \\
HetCache - slow                             & 50                & 30.68           & 176.31          &  $2.53\times$         & 16.50                   & 0.56                & 54.73                & 76.58              \\
HetCache - fast                             & 50                & \textbf{23.60}  & \textbf{166.81} &  $\mathbf{2.67}\times$& \textbf{16.58}          & \textbf{0.56}       & \textbf{54.51}       & 75.88              \\
{\textbf{Text-guided Video Editing on VPBench}}                 &                 &            &                  &           &                           &      &       &                      \\
{\textbf{Method}}                           & {\textbf{Step}}   & {\textbf{Efficiency}}  &       &           & {\textbf{Visual Quality}}   &      &       &                 \\
                                            &                   & \textbf{FLOPs (P) $\downarrow$}   & \textbf{Latency (s) $\downarrow$} & \textbf{Speed $\uparrow$} & \textbf{VFID $\downarrow$} & \textbf{LPIPS~$\uparrow$} & \textbf{VCLIP~$\uparrow$} & \textbf{VBench(\%)~$\uparrow$} \\
Wan2.1-VACE                                 & 75                & 64.59           & 246.05           & $1.00\times$  &  27.07                   & 0.29       & \textbf{0.30}     & 79.26             \\
Timestep Reduction                          & 50                & 43.06           & 174.03           & $1.41\times$  &  27.13                   & 0.29       & \textbf{0.30}     & 79.93            \\
TeaCache - slow                             & 50                & 27.99           & 163.52           & $1.50\times$  &  \textbf{25.64}          & \textbf{0.27}       & \textbf{0.30}     & \textbf{80.73}                   \\
TeaCache - fast                             & 50                & 21.53           & 137.70           & $1.79\times$  &  26.47                   & \textbf{0.27}       & \textbf{0.30}     & \textbf{80.73}              \\
HetCache - slow                             & 50                & 18.19           & 136.95           & $1.80\times$  &  26.89                   & \textbf{0.27}       & \textbf{0.30}     & 80.46                   \\
HetCache - fast                             & 50                & \textbf{13.99}  & \textbf{128.61}  & $\mathbf{1.91}\times$  & 27.14          & \textbf{0.27}       & \textbf{0.30}     & 80.59            
\end{tblr}
}
\vspace{-1em}
\end{table*}

\section{Experiments}\label{sec:experiments}

\subsection{Experiment Settings}

\noindent \textbf{Model Configurations.}
To evaluate the effectiveness of HetCache, we performed experiments in different video editing scenarios using Wan-2.1-VACE~\cite{wan2025wan}, one of the SOTA model with explicit support for VACE/MV2V tasks~\cite{jiang2025vace}. 
We primarily compare HetCache against TeaCache~\cite{liu2025timestep} which is well recognized as the state-of-the-art caching strategy for video diffusion models. 
In denoising timestep level, the “TeaCache-slow” and “TeaCache-fast" apply $\Delta$ equal to 0.05 and 0.02, respectively.
In our “HetCache-slow” and “HetCache-fast”, we set $\Delta$ to be 0.05 and 0.02, respectively, to ensure more intuitive comparison.
In spatio-temporal token level, both HetCache variants use identical token-selection hyper-parameters: $r_{\text{ctx}}=0.7$ (retain 70\% context tokens), $K=16$ (16 clusters in K-Means), and share the same $\alpha_i$ calculation. 

\begin{figure}[t]
    \includegraphics[width=\linewidth]{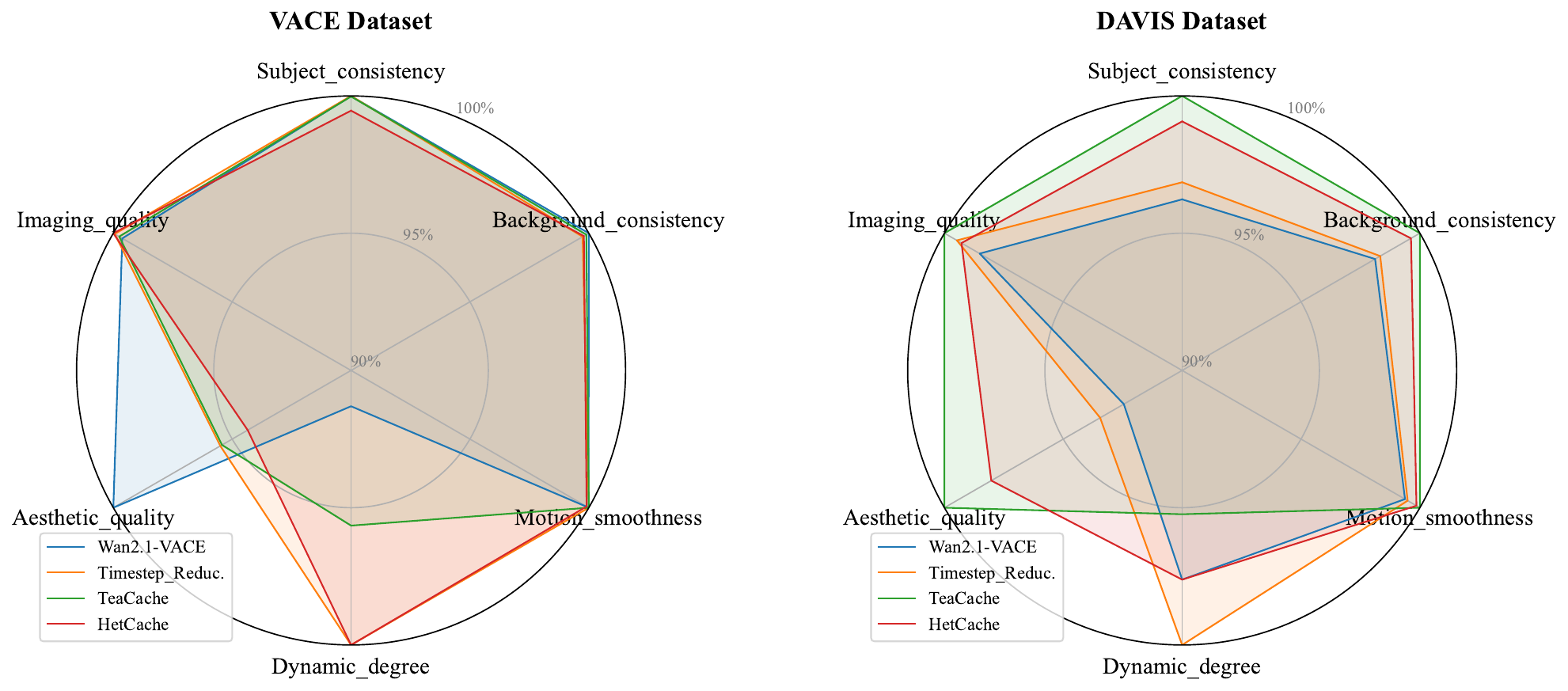}
    \caption{VBench comparison between HetCache and other methods on different video editing tasks.}
    \vspace{-1em}
    \label{fig:vbench-radar}
\end{figure}

\begin{figure*}[t]
    \centering\includegraphics[width=1\linewidth]{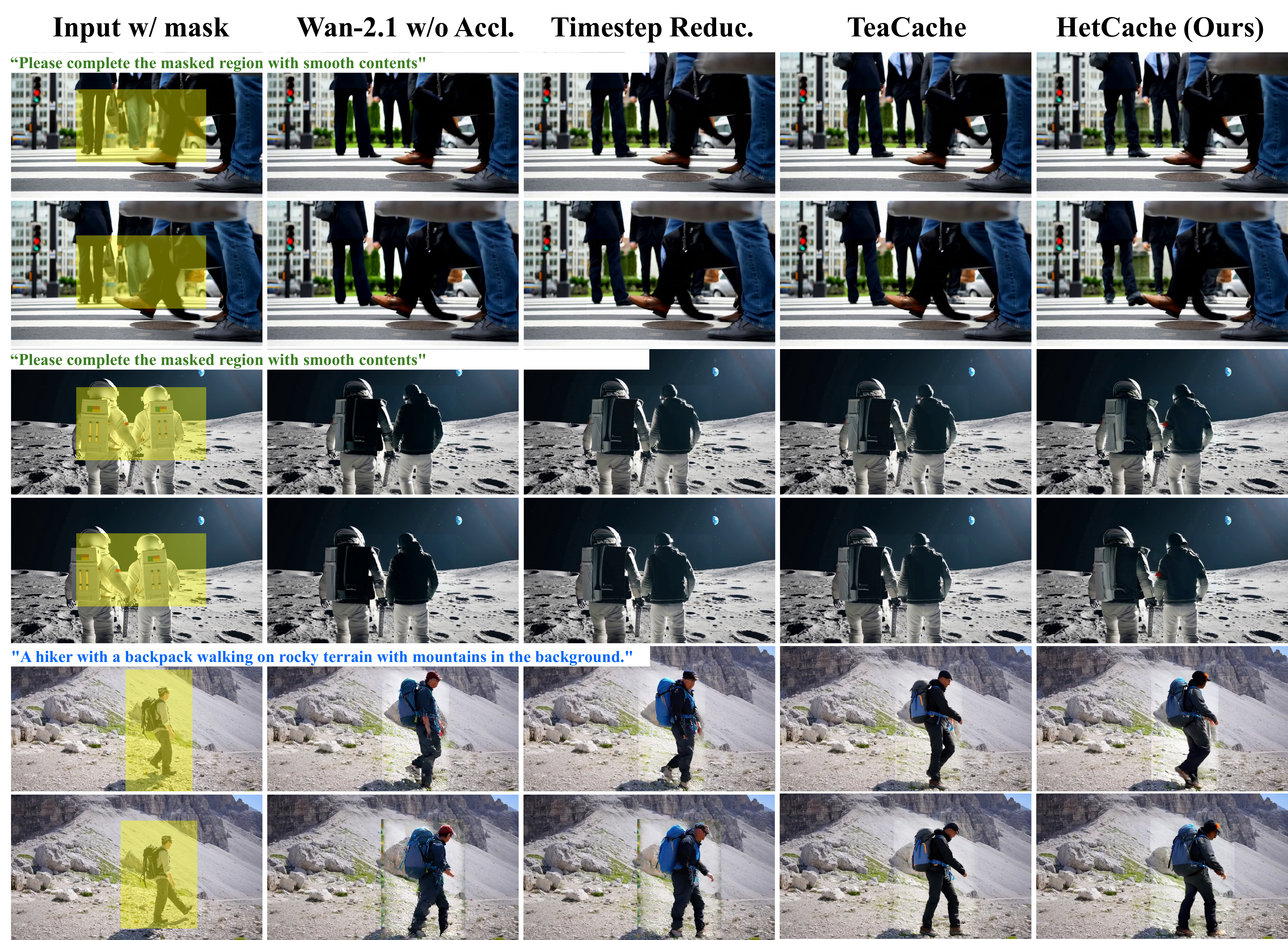}
    \caption{Visualization of different video editing tasks. HetCache produces relatively high-quality results while other methods suffer from smoothness, ghosting, and blurring issues.}
    \vspace{-1em}
    \label{fig:qual-eval}
\end{figure*}

\noindent \textbf{Evaluation and Metrics. } 
For MV2V-based video editing, we consider two common application scenarios: video inpainting/completion and text-guided partial video editing.
To evaluate inpainting quality, we use a sampled subset of the VACE-Benchmark~\cite{jiang2025vace}, measuring reconstruction fidelity with Peak Signal-to-Noise Ratio (PSNR), Structural Similarity Index Measure (SSIM), and Video Fréchet Inception Distance (VFID). 
In addition, we further assess inpainting performance on a DAVIS-derived~\cite{pont20172017} test set provided by VPBench~\cite{bian2025videopainter}.
For text-guided video generation, we focus on semantic alignment and perceptual quality, using VFID, LPIPS, and Video CLIP-score~\cite{wang2023exploring} as our main metrics. 
Beyond these task-specific metrics, both evaluation tracks also adopt the six-dimensional VBench evaluation protocol~\cite{huang2024vbench} to provide a comprehensive assessment of visual quality and temporal consistency.
Detailed experimental setting are provided in the supplementary materials.

\subsection{Quantitative Evaluation}
\noindent \textbf{Video Inpainting on VACE-Benchmark}
In VACE-Benchmark, HetCache consistently delivers the strongest computational savings among all methods. 
Compared with the 100-step Wan2.1-VACE full baseline (108.91 PFLOPs, 342.57 s), our HetCache-slow could approximately reduces compute to 30.68 PFLOPs and latency to 176.31 s under the same task setting, while HetCache-fast further brings FLOPs down to 23.60 PFLOPs and latency to 166.81 s, achieving up to a 2.67× speed-up. 
Importantly, these gains come with minimal quality impact, so PSNR/SSIM/VFID is outperformed. 
This indicates that, although HetCache accelerates more aggressively, it preserves the essential inpainting behavior and maintains stable visual quality.

Additionally, the VBench scores of the HetCache variants remain within a tight range around the baselines, as shown in Fig.~\ref{fig:vbench-radar}, the degradation in generation quality caused by HetCache is limited, but can help the model avoid some of the significant drawbacks of other methods, achieving a good balance across multiple dimensions with the lowest computational cost.

\noindent \textbf{Text-guided Video Editing on VPBench.}
A similar trend is observed on VPBench. 
HetCache achieves the lowest computation, theoretically 18.19 PFLOPs for HetCache-slow and 13.99 PFLOPs for HetCache-fast, corresponding to 1.9× acceleration over the 75-step baseline, while still keeping latency in a favorable range (136.95–128.61 s). 
Despite the reduction in FLOPs, HetCache maintains competitive visual quality. 
With all variants achieving VBench-Edit scores around 80\% and VFID, LPIPS, and VCLIP remaining aligned with the baseline, HetCache provides a reasonable efficiency-quality balance that maximizes computational reduction while maintaining editing fidelity.

We further evaluate HetCache in more configuration and task settings, as shown in Fig.~\ref{tab:more_accl}, we tested HetCache against TeaCache in higher resolutions, longer videos, outpainting tasks, and on an additional LTX~\cite{hacohen2024ltx} backbone, and the results showed a similar trend.

\begin{table}[h]
    \centering
    \scriptsize
    \setlength{\tabcolsep}{2.8pt}
    \renewcommand{\arraystretch}{1.03}
    \caption{Additional evaluation results under different settings.}
    \vspace{-0.5em}
    \label{tab:more_accl}
    \resizebox{\linewidth}{!}{
    \begin{tabular}{p{2.45cm}ccccccc}
        \toprule
        Method & FLOPs $\downarrow$ & Lat. $\downarrow$ & Spd. $\uparrow$ & PSNR $\uparrow$ & SSIM $\uparrow$ & VFID $\downarrow$ & VB $\uparrow$ \\
        \midrule
        \multicolumn{8}{l}{\textbf{Higher-resolution (25 × 720P) Video Inpainting on VACE-Benchmark }} \\
        Wan-VACE-1.3B (100) & 252.97 & 662.49 & 1.00$\times$ & 12.61 & 0.40 & 71.04 & 74.94 \\
        TeaCache (50)       & 126.48 & 276.05 & 2.40$\times$ & 13.01 & 0.41 & 73.71 & 75.03 \\
        HetCache (50)       & \textbf{107.89} & \textbf{227.54} & \textbf{2.91$\times$} & \textbf{13.03} & \textbf{0.41} & \textbf{71.40} & \textbf{75.68} \\
        \midrule
        \multicolumn{8}{l}{\textbf{Longer (57 × 480P) Video Inpainting on VACE-Benchmar}} \\
        Wan-VACE-1.3B (100) & 421.00 & 892.66 & 1.00$\times$ & 16.41 & 0.51 & 50.33 & 75.84 \\
        TeaCache (50)       & 210.50 & 364.82 & 2.44$\times$ & 17.09 & 0.52 & 48.86 & \textbf{76.81} \\
        HetCache (50)       & \textbf{179.56} & \textbf{291.46} & \textbf{3.06$\times$} & \textbf{17.12} & \textbf{0.52} & \textbf{47.19} & 75.85 \\
        \midrule
        \multicolumn{8}{l}{\textbf{Video Outpainting on VACE-Benchmark}} \\
        Wan-VACE-1.3B (100) & 154.93 & -- & -- & 19.49 & 0.62 & 43.44 & 76.56 \\
        TeaCache (50)       & 77.47 & -- & -- & 19.50 & 0.62 & \textbf{43.57} & 76.72 \\
        HetCache (50)       & \textbf{68.60} & -- & -- & \textbf{19.62} & \textbf{0.62} & 43.91 & \textbf{76.75} \\
        \midrule
        \multicolumn{8}{l}{\textbf{LTX-Video-VACE-based Video Inpainting on VACE-Benchmark}} \\
        LTX-Video-VACE-2B-0.9 (70) & -- & 140 & 1.00$\times$ & 14.72 & 0.58 & 64.50 & 80.41 \\
        TeaCache (70)              & -- & 133 & 1.05$\times$ & --    & --   & --    & --    \\
        HetCache (70)              & -- & 70  & 2.00$\times$ & 15.28 & 0.59 & 67.00 & 81.00 \\
        \bottomrule
    \end{tabular}
    }
    \vspace{-1em}
\end{table}

\subsection{Qualitative Evaluation}
In the visual comparison, we can see that in the scenario of masked video completion and generative editing, especially in the editing example of people hiking, HetCache not only has faster inference latency and lower computational cost, but also effectively prevents ghosting and dynamic boundary unsmoothness issues. 
In the static mask completion task, HetCache can also bring more details.

\subsection{Ablation Study}
\begin{figure}[t]
    \centering\includegraphics[width=\linewidth]{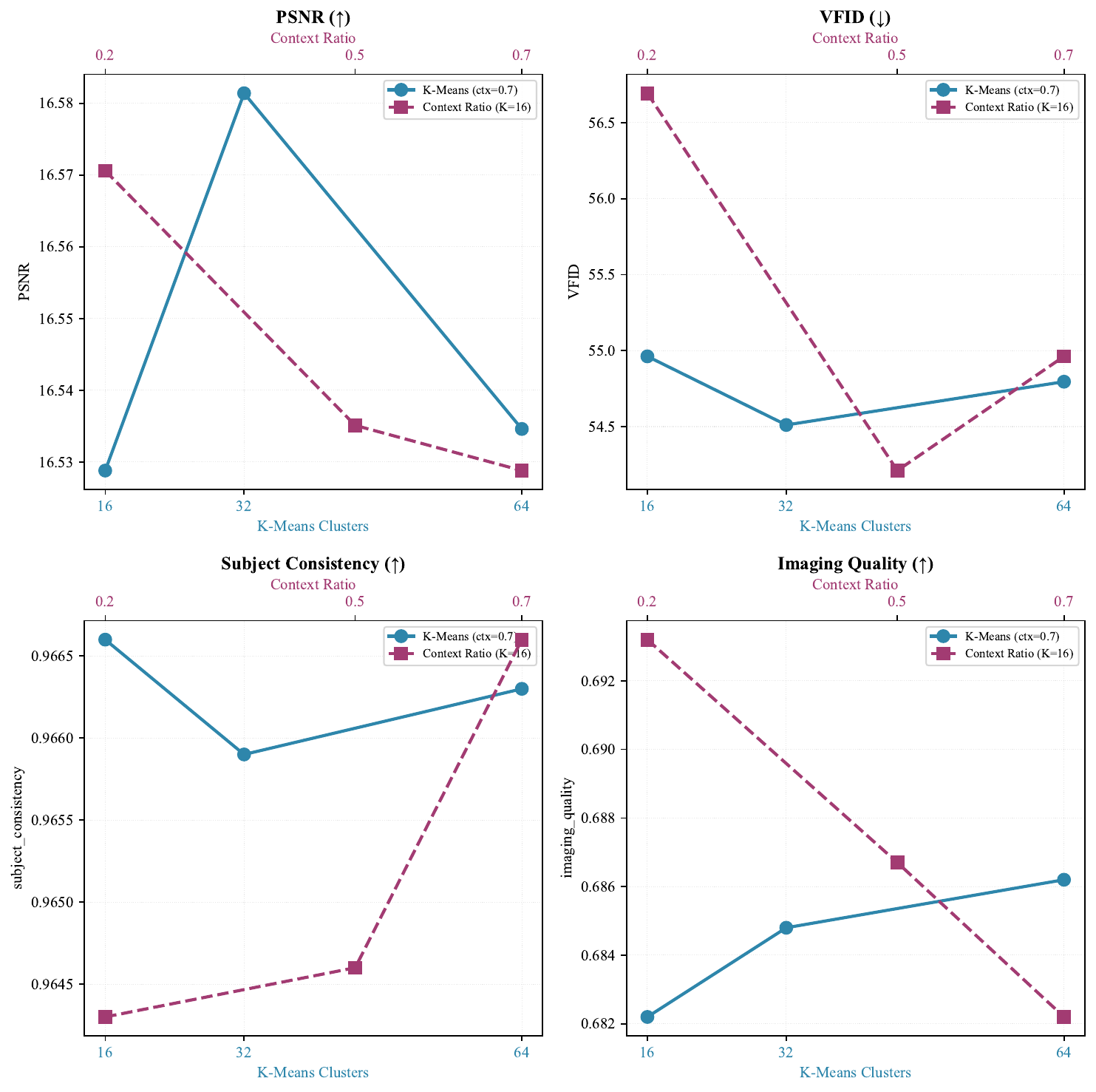}
    \centering\caption{Key metircs comparison of different $K$ and $r_{\text{ctx}}$ setting in context token sampling.}
    \label{fig:abl-clus-K}
    \vspace{-1em}
\end{figure}

\begin{figure}[t]
    \centering\includegraphics[width=\linewidth]{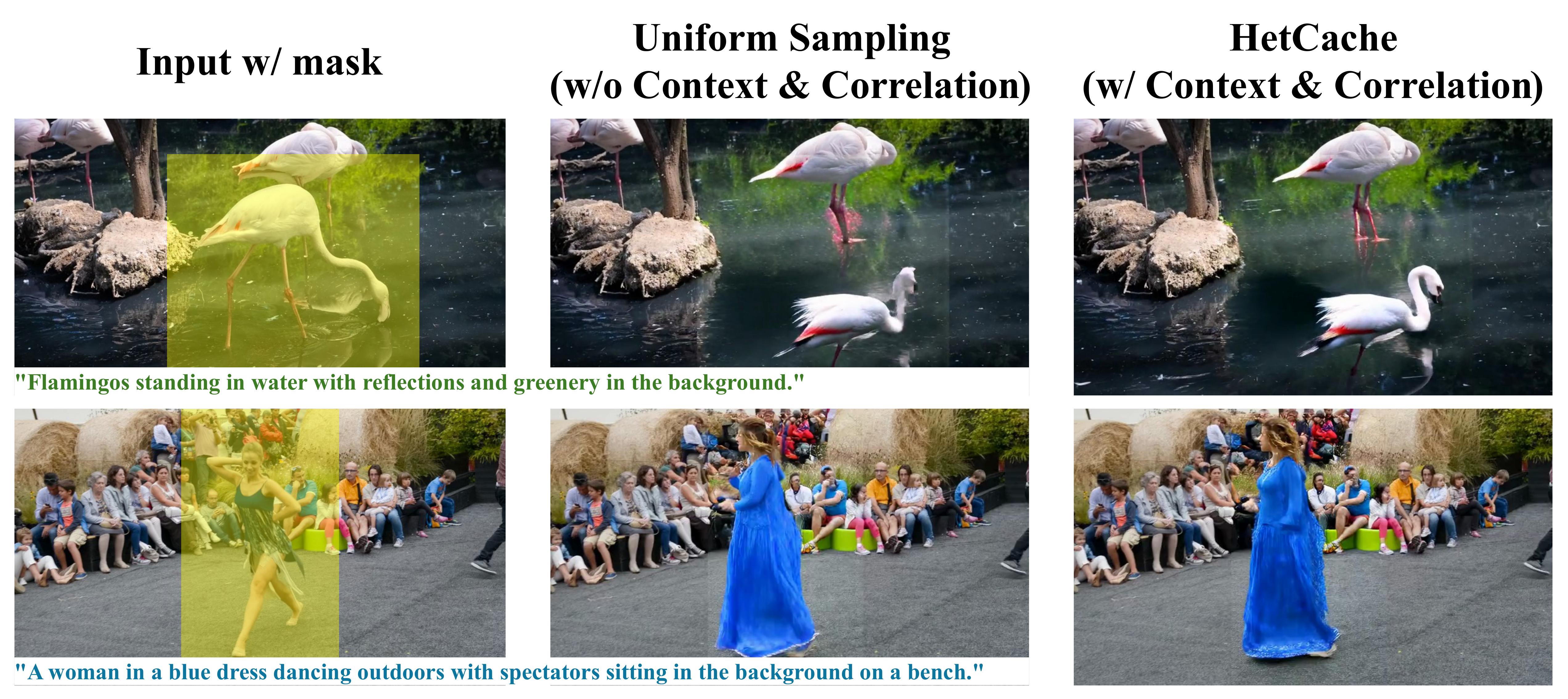}
    \caption{Visualization of ablation study, with and without clustering and correlation guidance will impact the generation quality.}
    \vspace{-1em}
    \label{fig:abl-vis}
\end{figure}

Our ablation study focused on the effectiveness of our token-level caching strategy components. 
Table.~\ref{tab:abmain} shows that when both the K-Means-based context representativeness and the sparse attention score-based correlation are discarded, uniform context token sampling (HetCache --) incurs a performance penalty, visualized in Fig.~\ref{fig:abl-vis}. 
Lower-quality context tokens directly reduce the generated quality of the target region, consistent with the inherent characteristics of editing characters. 
Furthermore, Fig. ~\ref{fig:abl-clus-K} shows that the selection of $K$ and context token parameters also leads to different performance impacts. 
Overall, keeping more context tokens generally leads to more robust performance, as expected.
Meanwhile, varying $K$ does not produce a monotonic trend which indicates that the semantic structure of context tokens has an effective capacity and does not benefit from arbitrarily fine partitioning.

\begin{table}[h]
\centering
\vspace{-0.5em}
\caption{Quantitative ablation study results.}
\vspace{-0.5em}
\label{tab:abl_efficiency}
\resizebox{\linewidth}{!}{
\begin{tblr}{
  column{2-9} = {c},
  cell{1}{1} = {r=2}{},
  cell{1}{2} = {c=2}{},
  cell{1}{4} = {c=2}{},
  cell{1}{6} = {c=4}{},
  vline{2,4,6} = {-}{},
  hline{1,7} = {-}{0.2em},
  hline{2,3} = {-}{},
}
\textbf{Method}       & \textbf{Guidance}    &                     & \textbf{Efficiency}           &                                      & \textbf{Visual Quality}                            &                                   &                                                      &                                           \\
                      & \textbf{Context} & \textbf{Correlation} & \textbf{Latency $\downarrow$} & \textbf{Speed $\uparrow$}~ & \textbf{PSNR$\uparrow$} & \textbf{SSIM~$\uparrow$} & \textbf{VFID$\downarrow$} & \textbf{VBench-Score~$\uparrow$} \\
\textbf{HetCache - -} & $\times$          & $\times$              & 142.46 & $3.13\times$ & \textbf{16.60}                         & \textbf{0.56}        & 54.54   & 76.19                                          \\
\textbf{HetCache -}   & $\checkmark$      & $\times$              & 152.14 & $2.93\times$ & 16.54                                  & \textbf{0.56}        & 54.75   & 75.80                                     \\
\textbf{HetCache -}   & $\times$          & $\checkmark$          & 177.46 & $2.51\times$ & \textbf{16.60}                         & \textbf{0.56}        & 55.36   & 76.24                                     \\
\textbf{HetCache}     & $\checkmark$      & $\checkmark$          & 166.81 & $2.67\times$ & 16.58                                  & \textbf{0.56}        & \textbf{54.51}   & \textbf{76.29}                                          
\end{tblr}
}
\vspace{-1em}
\label{tab:abmain}
\end{table}

\section{Conclusion}\label{sec:conclusion}
In this work, we presented HetCache, a training-free acceleration framework that leverages the inherent heterogeneity in diffusion-based video editing. By jointly exploiting variation across denoising timesteps and semantic correlation among spatio-temporal tokens, 
HetCache introduces heterogeneous caching that adaptively switches between full, partial, and reuse computation while selectively preserving informative context tokens. 
This design effectively reduces redundant attention operations and mitigates error accumulation during long denoising trajectories. 
Extensive experiments on VACE-Benchmark and VPBench demonstrate that HetCache achieves competitive visual quality with up to 2.67× speedup and significant FLOPs reduction, providing enhanced balance between efficiency and editing fidelity. 
We believe HetCache provides new insights into leveraging multidimensional redundancy for future Diffusion Transformer acceleration.

{
    \small
    \bibliographystyle{ieeenat_fullname}
    \bibliography{main}
}


\end{document}